\documentclass[11pt]{article}
\usepackage[a4paper,margin=2.3cm]{geometry}
\usepackage[T1]{fontenc}
\usepackage[utf8]{inputenc}
\usepackage{mathptmx}
\usepackage{microtype}
\usepackage{amsmath,amssymb}
\usepackage{booktabs}
\usepackage{array}
\newcolumntype{L}[1]{>{\raggedright\arraybackslash}p{#1}}
\usepackage{graphicx}
\usepackage{caption}
\usepackage{xcolor}
\usepackage[numbers,sort&compress]{natbib}
\usepackage[colorlinks=true,linkcolor=blue!50!black,citecolor=blue!50!black,urlcolor=blue!50!black]{hyperref}
\newcommand{\Hz}{\,Hz}

\title{History, Adaptation, and Hidden Gain:\\ Two Prospective Tests of Learned Predictive Control\\ in a Synthetic Neural Population}
\author{Mahesh Reddy Pagadala\\ \small Aston University, Birmingham, UK\\ \small \texttt{mahesh15116@gmail.com}}
\date{October 2026}

\begin{document}
\maketitle

\begin{abstract}
We study learned one-step predictive control of a synthetic population of 64 leaky integrate-and-fire neurons whose light sensitivity is scaled by a hidden factor $s$, using two prospective confirmations on plants simulated only after the analysis was frozen. In the first (50 plants, 11 conditions), removing the 200-ms input--output history substantially worsened a parameter-count-matched neural predictor at low and moderate gain ($s=0.6$, $0.8$; tracking RMSE rose by 2.47\Hz{} at $s=0.6$, 97.5\% CI 2.43--2.51); a fixed linear model with the same history was practically equivalent at low gain; in secondary and benchmark comparisons, recursively adapted ARX had the lowest mean error of the deployable controllers in 10 of 11 conditions, development-tuned PI with anti-windup matched or beat the neural predictor for $s\ge0.7$, and training the predictor for 10\,000 rather than 900 epochs lowered prediction error but worsened control. Development work then suggested one cause: under the tested conditions, the network's learned action sensitivity did not adequately track the hidden gain. A second protocol (50 new plants, 9 sensitivities, 2 noise levels) confirmed three pre-specified predictions at the primary noise level. Re-estimating a single multiplicative input gain recovered 87\% (95\% CI 86--88\%) of the benefit of full adaptive ARX. Giving the network that online gain estimate as an extra input reduced high-gain tracking error by 1.84\Hz{} [1.81, 1.88]. Longer training worsened the plain network by 1.09\Hz{} but improved the gain-input network by 1.07\Hz{}, a differential of 2.16\Hz{} [2.13, 2.20]. In secondary comparisons, the gain-input network came within 0.06\Hz{} of adaptive ARX and beat anti-windup PI at every sensitivity. Under high process noise the effects shrank to about 0.04\Hz{} or less and gain-only adaptation recovered only 24\% of the adaptive-ARX benefit. In this synthetic plant, at the primary noise level ($\sigma=0.5$), explicit online estimation of one hidden parameter by a one-parameter RLS estimator, an established idea in learning-based control, accounts for most of the gap between a history-based neural predictor and full adaptive ARX, and for the mismatch between its prediction and control performance; this did not hold at $\sigma=2.0$.
\end{abstract}

\section{Introduction}
Closed-loop control of neural population firing rates is used to probe circuit dynamics and is a component of proposed neural-interface systems. In optogenetic settings the effective plant gain can vary with opsin expression, optical attenuation and excitability, while the controller observes only the rate it produces~\cite{newman2015}. Learned dynamics models combined with sampling-based action selection are an attractive way to handle such systems~\cite{nagabandi2018}, but whether their extra capacity helps, and whether improving their predictive accuracy improves control, are empirical questions. We study them in a synthetic plant only and make no biological claim.

The work proceeded in two stages, each ending in a frozen protocol and a single run on untouched plants. The first stage asked: (i) does a learned one-step predictor need temporal input--output history once training data and capacity are matched; (ii) does a linear predictor with the same information suffice; (iii) how do learned and linear predictors compare with properly tuned PI and with online linear adaptation; and (iv) does a more accurate one-step predictor yield a better controller? Its answers raised a further question: (v) \emph{why} does online linear adaptation beat the learned predictor, and why does longer training make the learned predictor a worse controller? The second stage tested one mechanistic explanation for both: the hidden shift is a single multiplicative input gain, adaptive ARX succeeds by re-estimating it, and the network's learned action sensitivity does not adequately track it.

\paragraph{Contributions.}
(1)~A parameter-count-matched history ablation, an aligned-pair chronology perturbation and a practical-equivalence test of linear sufficiency. (2)~A regime-stratified comparison against PI tuned on development plants, with and without anti-windup, and against recursively adapted ARX. (3)~A pre-specified secondary comparison showing prediction--control mismatch arising from training length, an instance of objective mismatch~\cite{lambert2020}, later tested as a primary hypothesis. (4)~A prospective test showing that, at the primary noise level, a one-parameter gain estimate accounts for most of the adaptive-ARX advantage, and that supplying it to the network removes both the high-gain failure and the training-length mismatch. (5)~A documented methodological history (Appendix~\ref{app:history}): an earlier phase used an under-tuned PI and an RLS whose updates were negligible; both were corrected on development plants before any fresh confirmation.

\section{Related work}
\paragraph{Closed-loop optogenetic and neural control.} PI-type optogenetic feedback control of population activity has been demonstrated experimentally~\cite{newman2015}; the light required varied widely across preparations, which motivates treating plant gain as unknown. Model-based designs using identified input--output models have been developed for in-vivo use~\cite{bolus2018,grosenick2015}, including LQR control with a parameter-adaptive Kalman filter that tracks a drifting disturbance; re-estimating the input matrix was suggested as future work~\cite{bolus2021}. Learned forecasting models with model predictive control have been applied to a conductance-based neuron model, where a simple proportional controller was sometimes better~\cite{fehrman2023}, and to spiking networks via latent manifolds, where MPC outperformed PID~\cite{fehrman2024}. None of these compares a history-conditioned learned predictor with adaptive linear and tuned PI controllers under hidden gain shifts.

\paragraph{Identification, adaptation and history.} ARX models with recursive least squares and certainty-equivalence adaptive control are classical tools~\cite{astrom1995adaptive,ljung1999}; PI tuning and integrator anti-windup are covered in~\cite{astrom1995pid}. In learning-based control, recent input--output history is used to infer hidden parameters implicitly~\cite{kumar2021rma} or explicitly, by feeding an online system-identification estimate to a universal policy~\cite{yu2017}, typically under domain randomisation~\cite{tobin2017}. Our gain-input intervention is an instance of the explicit approach; it is not a new method.

\paragraph{Learned dynamics, fair baselines and objective mismatch.} That one-step model accuracy and control performance can diverge was shown in model-based reinforcement learning~\cite{lambert2020} and argued through value equivalence~\cite{grimm2020}. Equalising information and tuning between learned and classical controllers can shrink reported gaps~\cite{kunapuli2025}, although in-context learned models have outperformed PI on nonlinear motor loads~\cite{jian2026}.

\paragraph{Novelty boundary.} None of these methods or phenomena is new; we do not claim to have discovered objective mismatch or explicit parameter estimation. To our knowledge, the contribution is their combination in neural-population rate control: prospectively confirmed comparisons under hidden gain shifts, and a prospective test that, at the primary noise level, a one-parameter gain estimate accounts for most of the advantage of adaptive linear control and for the prediction--control mismatch of the learned predictor in this setting.

\section{Methods}
\subsection{Plant}
The plant is a population of $N=64$ leaky integrate-and-fire-like neurons sharing a scalar light command $u\in[0,1]$. Each neuron has initial potential $V_i(0)\sim U(-65,-55)$, bias $b_i\sim U(7,12)$ and light sensitivity $\kappa_i\sim U(18,26)$, drawn from a generator seeded by the plant seed (model voltage units). Each 50-ms control step is integrated with 50 Euler sub-steps of 1\,ms. A phenomenological first-order gate $q$ (time constant 10\,ms) filters the command, and each non-refractory neuron's potential is updated; both updates are applied in every 1-ms sub-step, in this order:
\begin{align}
q &\leftarrow q + (u-q)/10,\\
V_i &\leftarrow V_i + \tfrac{1}{20}\big[-(V_i+65) + b_i + s\,\kappa_i\, q\big] + \sigma\xi,\qquad \xi\sim\mathcal N(0,1).
\end{align}
A neuron reaching $-50$ spikes, resets to $-65$ and is refractory for 2\,ms. The observed output is the smoothed population rate $y_t=0.5\,y_{t-1}+0.5\,\rho_t$, where $\rho_t$ is the instantaneous population rate over the step (Hz per neuron). The hidden multiplier $s$ scales every $\kappa_i$; $\sigma$ is per-sub-step process noise. No controller receives $s$ or $\sigma$. Each episode lasts 180 steps (9\,s) and tracks 12, 30, 20, 40, 15 and 35\Hz, each held for 30 steps. The gate is not an opsin kinetic model.

\subsection{Training data and predictors}
All learned and linear predictors were fitted to plants 2000--2099 (training) and 2100--2129 (validation), 300 steps each, with per-plant $\sigma\sim U(0.5,2)$, $s\sim U(0.6,1.4)$ and bias shift $\sim U(-2,2)$, and the light command resampled from $U(0,1)$ every four steps (open-loop excitation). Rates are scaled by $1/60$. Networks are $9\to64\to1$ tanh multilayer perceptrons (DR-H4) whose inputs are the four past rates, the four past actions and a candidate action, trained by full-batch Adam on squared one-step error (learning rate 0.006), five initialisations each, keeping the best-validation weights within the stated epoch budget. The memoryless control (Mem-141, $3\to141\to1$, 706 parameters, matched to DR-H4 in parameter count but with a wider hidden layer) sees only the current rate, the previous action and the candidate. The ARX prior is linear in the same ten regressors (candidate action, four rates, four past actions, constant), fitted by least squares on the training data, with coefficient covariance $\Sigma_0=\hat\sigma^2(Z^\top Z)^{+}$, where $Z$ is the training regressor matrix, $\hat\sigma^2$ the residual variance and $^{+}$ the pseudo-inverse.

\subsection{Controllers}
All predictive controllers apply the action minimising $(\hat y_{t+1}(u)-r_t)^2+2u^2+2(u-u_{t-1})^2$, where $r_t$ is the target rate, over 101 candidates in $[0,1]$; this is one-step grid-search predictive control, not receding-horizon MPC or reinforcement learning. Table~\ref{tab:arms} lists the arms of both protocols.

\begin{table}[!htb]\centering\small
\caption{Controllers. Network arms have five initialisations each, averaged within plant. Every gain, epoch and checkpoint was fixed on training, validation or development plants before the corresponding confirmation.}\label{tab:arms}
\begin{tabular}{@{}L{3.1cm}L{9.6cm}L{1.6cm}@{}}\toprule
Arm & Definition & Protocol\\\midrule
PI tuned & positional PI, $K_p=0.02$, $K_i=0.20$, integrator clip $\pm10$ (development grid of 400 gain pairs) & v3, v4\\
PI tuned + AW & as above with back-calculation anti-windup, $K_p=0.02$, $K_i=0.25$, $k_t=0.5$ & v3, v4\\
PI sens.\ ref. & best PI of the family when $s$ is known; not deployable, reference only & v3\\
ARX fixed & least-squares prior, no online update & v3, v4\\
ARX + RLS & all ten coefficients adapted by RLS, $P_0=10^9\,\Sigma_0$, $\lambda=0.98$ & v3, v4\\
ARX-$g$ & one-parameter gain RLS (Section~\ref{sec:gain}), $P_0=1000$, $\lambda=0.98$ & v4\\
H4-10k & DR-H4, 10\,000-epoch budget, best validation & v3, v4\\
H4-900 & DR-H4, 900-epoch budget & v3, v4\\
Mem-141 & memoryless network, 10\,000-epoch budget & v3\\
I1 & DR-H4 raw snapshot at a per-seed epoch (300--900) chosen by closed-loop error on development plants 5250--5299 & v4\\
I2 & DR-H4 with the online gain estimate $g$ as a tenth input; best validation within 6000 epochs & v4\\
I3 & DR-H4 retrained on open-loop data plus closed-loop trajectories of its own 10k model; best validation within 6000 epochs & v4\\
H4 / I2 at epoch 900, 6000 & raw weights at fixed epochs from checkpointed retraining (for the matched-epoch test) & v4\\\bottomrule
\end{tabular}\end{table}

\subsection{One-parameter gain estimator}\label{sec:gain}
Rather than re-estimating every coefficient of the linear dynamics model, ARX-$g$ retains the previously learned dynamics and updates a single multiplier governing the combined effect of the current and recent light commands. This multiplier is inferred from observed prediction errors during closed-loop operation. Let $z_t$ be the ten regressors of the ARX prior $w_0$ (candidate action, four past rates, four past actions, constant). Split the prior into action terms $a_t=z_t^\top w_a$ (candidate and four past actions) and the rest $z_t^\top w_r$. The gain estimator predicts $\hat y_{t+1}=z_t^\top w_r + g\,a_t$ and, after every step, updates the scalar $g$ by RLS:
\begin{align}
k_t &= \frac{P_t\,a_t}{\lambda + a_t^2 P_t},\nonumber\\
g_{t+1} &= g_t + k_t\,\big(y_{t+1} - z_t^\top w_r - g_t a_t\big),\\
P_{t+1} &= \frac{P_t - k_t\,a_t P_t}{\lambda},\nonumber
\end{align}
with $\lambda=0.98$, $P_0=1000$ and $g_0=1$, all rates in scaled units ($\div60$). Because the action terms include the current candidate and four past actions, $g$ represents an effective input-response scaling, not simply the instantaneous derivative of firing rate with respect to light. ARX-$g$ uses this predictor for control. In I2 the same estimator runs alongside the network, and its current $g$ is the network's tenth input; for training, $g$ was computed on the open-loop training trajectories before each next rate was observed. The estimator, its settings and its use are identical at both noise levels.

\subsection{Interventions and checkpoints}
Three interventions were developed on plants 5200--5299 to address the training-length mismatch: I1 (closed-loop early stopping), I2 (gain input) and I3 (on-policy data; closed-loop trajectories at random targets of 10--45\Hz{} held for 30 steps). For the matched-epoch test the plain DR-H4 was retrained with weights saved at fixed epochs; its epoch-900 weights reproduce the original 900-epoch weights to within $10^{-15}$. No checkpoint, epoch, model variant or estimator setting was chosen on confirmation plants.

\subsection{Study design}
Each protocol was frozen before its confirmation plants were instantiated, with SHA-256 hashes of the script, protocol, simulator modules, ARX prior and every weight file recorded in a manifest that the run script re-checks on every invocation. Each confirmation block was simulated as a single planned campaign, with no outcome-driven reruns, using the exact scalar simulator (Appendix~\ref{app:integrity} documents the infrastructure-driven recomputations this still required). Confirmation blocks were never reused, and no development block overlaps a confirmation block (Appendix~\ref{app:history}). This design supports prospective confirmation of pre-specified predictions on untouched plants within the same simulator and plant family; it is not independent replication across different neural models, architectures or biological systems, and we make no such claim.

\textbf{Protocol v3} (plants 7000--7049): $s\in\{0.5,0.6,\dots,1.2,1.4\}$ at $\sigma=0.5$ ($s=0.5$ lies outside the training range), plus $\sigma\in\{1,2\}$ at $s=1$. \textbf{Protocol v4} (plants 8000--8049): the same nine sensitivities at $\sigma=0.5$ (primary) and $\sigma=2.0$ (robustness), 900 plant-conditions.

\subsection{Hypotheses and statistics}
The unit of analysis is the plant: network RMSE (180 steps) is averaged over the five initialisations within each plant before paired differences are formed. Intervals are bias-corrected and accelerated (BCa) bootstrap intervals~\cite{efron1987} with 10\,000 resamples and a fixed bootstrap seed per interval, Bonferroni-corrected within each family. Table~\ref{tab:hyp} lists all pre-specified tests. In v4, the high-gain set $\mathcal H=\{1.0,1.1,1.2,1.4\}$ was fixed in advance, and Q1 is the ratio of plant-level means
\begin{equation}
Q_1=\frac{\overline{\textstyle\frac19\sum_s(\mathrm{ARX}_{\mathrm{fixed}}-\mathrm{ARX}_g)}}{\overline{\textstyle\frac19\sum_s(\mathrm{ARX}_{\mathrm{fixed}}-\mathrm{ARX}_{\mathrm{RLS}})}},
\end{equation}
with the bar denoting the mean over plants. Q4 is a difference-in-differences at matched epochs, $D=[\text{H4}_{6000}-\text{H4}_{900}]-[\text{I2}_{6000}-\text{I2}_{900}]$, averaged over $\mathcal H$. A positive $D$ supports a differential effect only; the directions of the two components were tested separately. A former hypothesis Q2 (monotonicity of $g$ in $s$) was moved to secondary before the freeze, hence the gap in numbering.

\begin{table}[t]\centering\small
\caption{Pre-specified tests. Secondary results do not change primary decisions. Development values for v4 came from plants 5200--5249.}\label{tab:hyp}
\begin{tabular}{@{}llL{7.6cm}L{4.4cm}@{}}\toprule
Protocol & ID & Contrast & Level / decision\\\midrule
v3 & P1 & Mem-141 $-$ H4-10k at $s=0.6, 0.8$ & 97.5\% each; both lower bounds $>0$\\
v3 & P2 & ARX fixed $-$ H4-10k at $s=0.6$ & 95\%; interval within $\pm0.25$\Hz\\
v3 & B1 & PI tuned (+AW) $-$ H4-10k, 11 conditions & 99.77\% each; no direction\\
v3 & S1, S3--S5 & RLS effect, chronology shuffle, ARX+RLS $-$ H4-10k, H4-900 $-$ H4-10k & 97.5--99.55\%; exploratory\\
v4 & Q1 & gain fraction, $\sigma=0.5$ & 95\%; lower bound $\ge0.70$\\
v4 & Q3 & H4-10k $-$ I2, mean over $\mathcal H$, $\sigma=0.5$ & 95\%; lower bound $>0$\\
v4 & Q4 & $D$ over $\mathcal H$, $\sigma=0.5$ & 95\%; lower bound $>0$\\
v4 & secondary & directions of Q4 components; Spearman$(g,s)$ per plant; calibration of $g$; Q3 per setting; I2 $-$ ARX+RLS (no equivalence claim); I1, I3 vs H4-10k; I2 $-$ PI+AW; action-sensitivity slope & 97.5--99.44\%\\
v4 & R & Q1, Q3, Q4 at $\sigma=2.0$ & 98.3\% each; robustness\\\bottomrule
\end{tabular}\end{table}

\section{Results: protocol v3}
\begin{figure}[t]\centering
\includegraphics[width=0.78\linewidth]{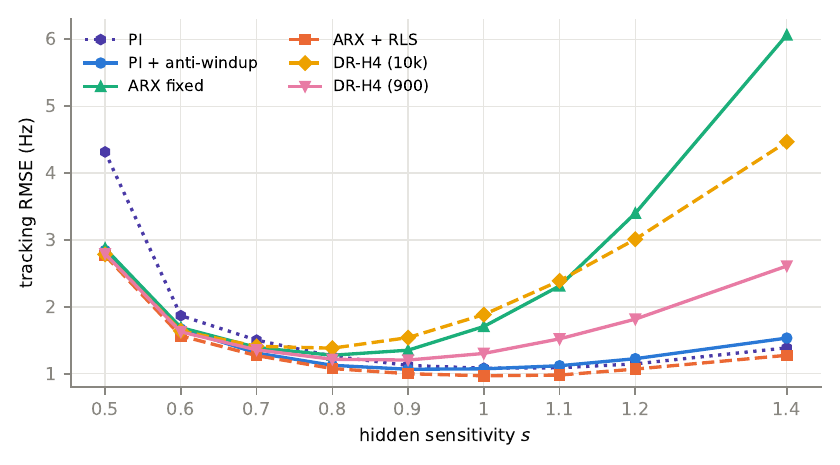}
\caption{Protocol v3: mean tracking RMSE against hidden sensitivity at $\sigma=0.5$. The memoryless network (1.7--5.6\Hz) is omitted for scale; all values are in Table~\ref{tab:v3}.}\label{fig:v3}
\end{figure}

Both primary hypotheses were supported. Table~\ref{tab:v3} and Figure~\ref{fig:v3} give mean tracking RMSE for every controller and condition.

\begin{table}[t]\centering\small
\caption{Protocol v3: mean tracking RMSE (Hz) over 50 confirmation plants (7000--7049). Lowest of the deployable arms shown in bold. $^\dagger$Uses knowledge of $s$; excluded from bold. A legacy PI with phase-1 gains ($K_i=0.15$, not shown) was 0.004\Hz{} lower than ARX+RLS at $s=1.4$ and higher elsewhere.}\label{tab:v3}
\begin{tabular}{@{}lrrrrrrrr@{}}\toprule
Condition & PI & PI+AW & ARX & ARX+RLS & H4-10k & H4-900 & Mem-141 & PI ref.$^\dagger$ \\\midrule
$s=0.50$ & 4.316 & 2.844 & 2.873 & \textbf{2.769} & 2.784 & 2.794 & 5.572 & 2.664 \\
$s=0.60$ & 1.870 & 1.670 & 1.692 & \textbf{1.567} & 1.642 & 1.627 & 4.111 & 1.471 \\
$s=0.70$ & 1.504 & 1.311 & 1.394 & \textbf{1.273} & 1.412 & 1.359 & 3.119 & 1.226 \\
$s=0.80$ & 1.264 & 1.128 & 1.278 & \textbf{1.080} & 1.384 & 1.218 & 2.425 & 1.124 \\
$s=0.90$ & 1.126 & 1.069 & 1.353 & \textbf{1.002} & 1.542 & 1.206 & 1.966 & 1.070 \\
$s=1.00$ & 1.084 & 1.075 & 1.707 & \textbf{0.971} & 1.886 & 1.305 & 1.748 & 1.075 \\
$s=1.10$ & 1.091 & 1.121 & 2.314 & \textbf{0.982} & 2.390 & 1.523 & 1.743 & 1.080 \\
$s=1.20$ & 1.149 & 1.227 & 3.400 & \textbf{1.071} & 3.012 & 1.818 & 1.909 & 1.045 \\
$s=1.40$ & 1.389 & 1.535 & 6.063 & \textbf{1.277} & 4.467 & 2.611 & 2.523 & 1.067 \\
$\sigma=1.0$ & 1.311 & 1.271 & 1.339 & \textbf{1.204} & 1.324 & 1.277 & 1.589 & n/a \\
$\sigma=2.0$ & 3.231 & 2.261 & 2.204 & \textbf{2.136} & 2.158 & 2.163 & 2.965 & n/a \\
\bottomrule\end{tabular}\end{table}

\paragraph{History substantially helps at low and moderate gain (P1).} With training data and parameter count matched, removing the history substantially worsened the trained predictor: RMSE increased by +2.47 [+2.43, +2.51]\Hz{} at $s=0.6$ and +1.04 [+1.01, +1.08]\Hz{} at $s=0.8$ (97.5\% intervals), on every plant. In a secondary test (S3), a joint permutation of the history raised RMSE by +3.68 [+3.47, +3.88] and +6.75 [+6.49, +6.99]\Hz{}. Because permuted histories never occur in training, this shows that the controller depends on temporal order, not that it estimates the plant gain. P1 was tested only at $s=0.6$ and $0.8$, so no universal claim about history follows; at $s\ge1.0$ the memoryless network tracked better than H4-10k (Table~\ref{tab:v3}), consistent with the 10k network's high-gain failure examined in Section~6.

\paragraph{A linear model with the same history suffices at low gain (P2).} At $s=0.6$ the fixed ARX differed from H4-10k by +0.050 [+0.038, +0.063]\Hz{} (95\%), within the $\pm0.25$-Hz margin. Away from low gain the fixed ARX degrades sharply (3.40\Hz{} at $s=1.2$, 6.06\Hz{} at 1.4).

\paragraph{Tuned PI matches or beats the learned predictor from $s=0.7$ (B1).} The network kept an advantage only where PI was saturated or noise-limited: at $s\le0.6$, and at $\sigma=2$, where the zero-light rate floor exceeds the 12-Hz target (anti-windup PI +0.10\Hz). Against anti-windup PI the low-gain advantage was small (+0.03\Hz{} at $s=0.6$). From $s=0.7$ upwards anti-windup PI was better on every plant, by 0.10\Hz{} at 0.7 rising to 2.93\Hz{} at 1.4 (Table~\ref{tab:b1}).

\paragraph{Online linear adaptation had the lowest error (S1, S4).} RLS improved the fixed ARX in every condition, by 0.07\Hz{} at $\sigma=2$ to 4.79\Hz{} at $s=1.4$. Adaptive ARX had a lower mean RMSE than H4-10k in all 11 conditions; the paired 99.55\% interval excluded zero in 9 of them, but not at $s=0.5$ ($-$0.015 [$-$0.038, +0.010]\Hz) or $\sigma=2$ ($-$0.023 [$-$0.050, +0.004]\Hz). Among the controllers displayed in Table~\ref{tab:v3}, adaptive ARX achieved the lowest mean tracking RMSE in all 11 conditions; an additional legacy PI configuration, retained in the study records but omitted from the table, achieved an RMSE lower by 0.004\Hz{} at $s=1.4$ (Table~\ref{tab:v3} caption). Adaptive ARX also came within 0.03\Hz{} of the sensitivity-informed PI reference at $s=1.2$ (1.071 vs 1.045\Hz).

\paragraph{Better prediction, worse control (S5, secondary).} The 10k models had lower one-step validation RMSE than the 900-epoch models (1.39--1.40 vs 1.61--1.85\Hz) but tracked worse at every $s\ge0.6$ and at $\sigma=1$ (by only 0.015\Hz{} at $s=0.6$): by 0.05\Hz{} at $s=0.7$, 0.58\Hz{} at 1.0 and 1.86\Hz{} at 1.4. This comparison was added after a development observation and was secondary.

\begin{table}[t]\centering\small
\caption{Protocol v3: PI minus H4-10k RMSE (Hz) with simultaneous 95\% intervals (99.77\% each). Negative values favour PI.}\label{tab:b1}
\begin{tabular}{@{}lll@{}}\toprule
Condition & PI tuned $-$ H4-10k & PI + AW $-$ H4-10k\\\midrule
$s=0.50$ & +1.532 [+1.485, +1.569] & +0.060 [+0.045, +0.076] \\
$s=0.60$ & +0.228 [+0.209, +0.248] & +0.028 [+0.008, +0.046] \\
$s=0.70$ & +0.092 [+0.070, +0.110] & $-$0.101 [$-$0.122, $-$0.082] \\
$s=0.80$ & $-$0.119 [$-$0.146, $-$0.097] & $-$0.256 [$-$0.285, $-$0.233] \\
$s=0.90$ & $-$0.416 [$-$0.456, $-$0.379] & $-$0.473 [$-$0.511, $-$0.431] \\
$s=1.00$ & $-$0.801 [$-$0.851, $-$0.759] & $-$0.810 [$-$0.860, $-$0.762] \\
$s=1.10$ & $-$1.299 [$-$1.353, $-$1.247] & $-$1.269 [$-$1.317, $-$1.223] \\
$s=1.20$ & $-$1.862 [$-$1.930, $-$1.800] & $-$1.785 [$-$1.854, $-$1.720] \\
$s=1.40$ & $-$3.078 [$-$3.147, $-$2.993] & $-$2.932 [$-$3.003, $-$2.848] \\
$\sigma=1.0$ & $-$0.013 [$-$0.038, +0.014] & $-$0.053 [$-$0.071, $-$0.034] \\
$\sigma=2.0$ & +1.073 [+1.010, +1.129] & +0.103 [+0.073, +0.129] \\
\bottomrule\end{tabular}\end{table}

\section{From v3 to v4: development findings}
On fresh development plants (5200--5249 for evaluation; 5250--5299 only for selecting I1's epochs), three observations motivated the second protocol. (a)~In the full RLS, the coefficient on the candidate action tracked $s$ almost perfectly, but adapting that coefficient alone recovered little, because the gain also enters through the four lagged-action terms. One scalar multiplying all five action terms tracked $s$ linearly and recovered 86\% of the full-RLS benefit. (b)~Across 60 training checkpoints, the correlation between open-loop validation error and closed-loop error fell from 0.81 at $s=0.6$ to $-0.08$ at $s=1.4$; the network's local action sensitivity $\partial\hat y/\partial u$ stayed nearly constant across $s$ instead of scaling with the plant gain. (c)~Of three interventions, supplying the gain estimate as an input (I2) brought the network's mean RMSE from 2.29 to 1.35\Hz, close to full adaptive ARX (1.33\Hz), and made longer training help instead of hurt. All three findings were then frozen as v4 hypotheses (Table~\ref{tab:hyp}).

\section{Results: protocol v4}
\begin{figure}[t]\centering
\includegraphics[width=\linewidth]{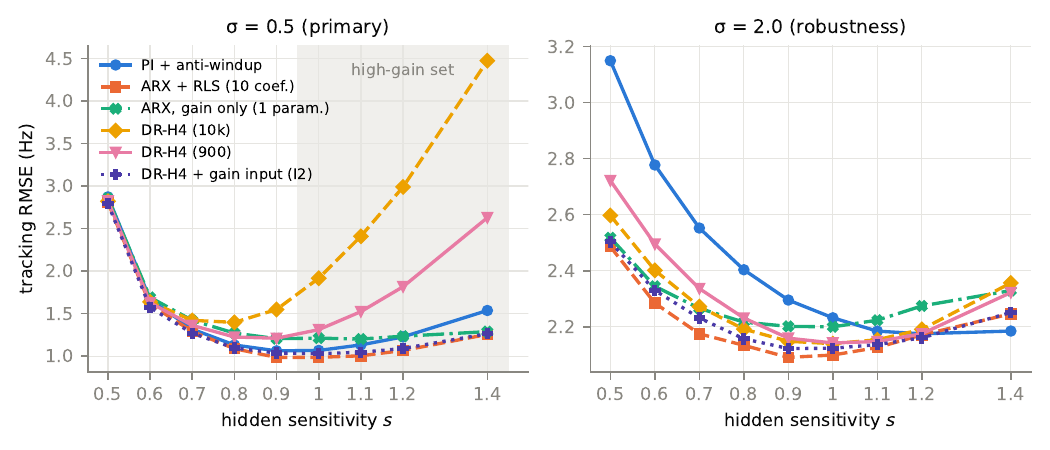}
\caption{Protocol v4: mean tracking RMSE against hidden sensitivity for six arms. Shading marks the pre-specified high-gain set. All arms are in Table~\ref{tab:v4}.}\label{fig:v4}
\end{figure}

All three primary hypotheses were supported (Table~\ref{tab:primary}). Mean RMSE for every arm is in Table~\ref{tab:v4} and Figure~\ref{fig:v4}.

\begin{table}[t]\centering\small
\caption{Protocol v4 primary decisions at $\sigma=0.5$ (plants 8000--8049; 95\% BCa intervals) and the same statistics at $\sigma=2.0$ (98.3\% intervals; robustness, not decisions). $^*$Development values at $s=1.0$ and $1.4$; no high-gain-set mean was recorded.}\label{tab:primary}
\begin{tabular}{@{}llllll@{}}\toprule
ID & Criterion & Development & $\sigma=0.5$ & Decision & $\sigma=2.0$\\\midrule
Q1 & lower bound $\ge0.70$ & 0.86 & 0.871 [0.863, 0.878] & supported & 0.24 [0.17, 0.32]\\
Q3 & lower bound $>0$ & +1.84\Hz & +1.84 [1.81, 1.88] & supported & +0.042 [0.033, 0.054]\\
Q4 & lower bound $>0$ & +1.0 to +3.8\Hz$^*$ & +2.16 [2.13, 2.20] & supported & +0.018 [0.008, 0.027]\\\bottomrule
\end{tabular}\end{table}

\begin{table}[t]\centering\small
\caption{Protocol v4: mean tracking RMSE (Hz) over 50 plants. ARX-$g$: one-parameter gain RLS; I1--I3: interventions. Lowest in each row in bold. H4 and I2 at fixed epochs are in Figure~\ref{fig:train}.}\label{tab:v4}
\begin{tabular}{@{}lrrrrrrrrr@{}}\toprule
\multicolumn{10}{@{}l}{\emph{$\sigma=0.5$ (primary)}}\\
$s$ & PI+AW & ARX & ARX+RLS & ARX-$g$ & H4-10k & H4-900 & I1 & I3 & I2 \\\midrule
0.50 & 2.872 & 2.901 & 2.806 & 2.857 & 2.822 & 2.825 & 2.851 & 2.808 & \textbf{2.792} \\
0.60 & 1.668 & 1.696 & 1.608 & 1.691 & 1.638 & 1.627 & 1.647 & 1.595 & \textbf{1.571} \\
0.70 & 1.317 & 1.400 & 1.290 & 1.421 & 1.421 & 1.362 & 1.362 & 1.321 & \textbf{1.269} \\
0.80 & 1.131 & 1.261 & \textbf{1.088} & 1.271 & 1.397 & 1.225 & 1.206 & 1.179 & 1.095 \\
0.90 & 1.062 & 1.367 & \textbf{0.985} & 1.206 & 1.549 & 1.210 & 1.158 & 1.160 & 1.037 \\
1.00 & 1.067 & 1.695 & \textbf{0.986} & 1.212 & 1.916 & 1.312 & 1.204 & 1.259 & 1.029 \\
1.10 & 1.132 & 2.326 & \textbf{1.004} & 1.201 & 2.410 & 1.523 & 1.337 & 1.437 & 1.046 \\
1.20 & 1.226 & 3.378 & \textbf{1.066} & 1.234 & 2.990 & 1.816 & 1.516 & 1.679 & 1.090 \\
1.40 & 1.537 & 6.053 & \textbf{1.255} & 1.288 & 4.474 & 2.628 & 2.025 & 2.262 & 1.264 \\
\midrule mean & 1.446 & 2.453 & \textbf{1.343} & 1.487 & 2.291 & 1.725 & 1.590 & 1.633 & 1.355 \\
\midrule
\multicolumn{10}{@{}l}{\emph{$\sigma=2.0$ (robustness)}}\\
$s$ & PI+AW & ARX & ARX+RLS & ARX-$g$ & H4-10k & H4-900 & I1 & I3 & I2 \\\midrule
0.50 & 3.149 & 2.646 & \textbf{2.486} & 2.518 & 2.597 & 2.721 & 2.824 & 2.673 & 2.503 \\
0.60 & 2.777 & 2.411 & \textbf{2.284} & 2.344 & 2.401 & 2.495 & 2.566 & 2.472 & 2.329 \\
0.70 & 2.552 & 2.284 & \textbf{2.175} & 2.265 & 2.272 & 2.336 & 2.383 & 2.334 & 2.231 \\
0.80 & 2.403 & 2.222 & \textbf{2.134} & 2.217 & 2.192 & 2.231 & 2.261 & 2.224 & 2.157 \\
0.90 & 2.295 & 2.183 & \textbf{2.091} & 2.201 & 2.148 & 2.159 & 2.188 & 2.167 & 2.122 \\
1.00 & 2.232 & 2.161 & \textbf{2.099} & 2.200 & 2.140 & 2.142 & 2.150 & 2.120 & 2.123 \\
1.10 & 2.184 & 2.206 & \textbf{2.126} & 2.223 & 2.154 & 2.147 & 2.140 & 2.126 & 2.136 \\
1.20 & 2.175 & 2.271 & 2.171 & 2.274 & 2.191 & 2.175 & 2.166 & \textbf{2.143} & 2.161 \\
1.40 & \textbf{2.185} & 2.432 & 2.246 & 2.329 & 2.356 & 2.322 & 2.297 & 2.269 & 2.251 \\
\midrule mean & 2.439 & 2.313 & \textbf{2.201} & 2.286 & 2.272 & 2.303 & 2.331 & 2.281 & 2.224 \\
\bottomrule\end{tabular}\end{table}

\subsection{At \texorpdfstring{$\sigma=0.5$}{sigma=0.5}, one input gain accounts for most of the adaptive-ARX advantage (Q1)}
\begin{figure}[t]\centering
\includegraphics[width=0.45\linewidth]{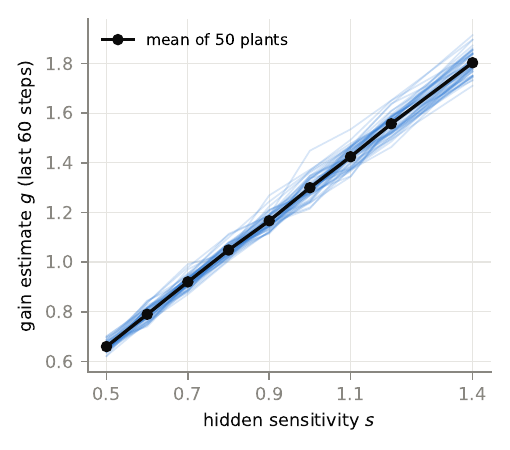}
\caption{Gain estimate $g$ (mean of the last 60 steps of ARX-$g$) against $s$ for each of the 50 v4 plants (thin lines) and their mean, $\sigma=0.5$.}\label{fig:g}
\end{figure}

Adapting the single scalar $g$ recovered 0.871 [0.863, 0.878] of the improvement that full ten-coefficient RLS achieves over the fixed ARX, above the 0.70 threshold. In secondary analyses, the estimate behaved as a gain estimate should (Figure~\ref{fig:g}): its rank correlation with $s$ was 1.0 on every one of the 50 plants (minimum 1.00); a per-plant linear fit gave slope 1.27 [1.26, 1.29] (99\%), intercept 0.028 and mean absolute residual 0.017; the coefficient of variation of $g/s$ was 0.021. Because $g$ multiplies the domain-randomised prior's action terms, its absolute scale is relative to that prior ($g=1$ at $s=1$ would mean no correction; observed $g-1=$ 0.30).

\subsection{Supplying the gain estimate removes the high-gain failure (Q3)}
Over the high-gain set, I2 tracked better than H4-10k by 1.84\Hz{} [1.81, 1.88]. The advantage held at every high-gain setting and on every plant (Table~\ref{tab:q3}). Across all nine sensitivities I2 had a mean RMSE of 1.355\Hz, against 2.291\Hz{} for H4-10k and 1.343\Hz{} for full adaptive ARX. In a secondary analysis, the learned action sensitivity now scaled with the plant gain: the slope of $\partial\hat y/\partial u$ against $s$, relative to the slope of the gain-RLS reference, was 0.84 [0.83, 0.85] for I2 versus 0.39 [0.39, 0.40] for H4-10k and 0.09 [0.09, 0.09] for H4-900. The reference is itself the one-parameter estimate, not the plant's true local gain.

\begin{table}[t]\centering\small
\caption{H4-10k $-$ I2 at each high-gain setting ($\sigma=0.5$; 98.75\% intervals) and the share of plants with a positive difference.}\label{tab:q3}
\begin{tabular}{@{}lll@{}}\toprule
$s$ & H4-10k $-$ I2 (Hz) & plants $>0$\\\midrule
1.00 & +0.89 [+0.85, +0.93] & 100\% \\
1.10 & +1.36 [+1.32, +1.41] & 100\% \\
1.20 & +1.90 [+1.84, +1.97] & 100\% \\
1.40 & +3.21 [+3.15, +3.28] & 100\% \\
\bottomrule\end{tabular}\end{table}

\subsection{The training-length mismatch is specific to the network without the gain (Q4)}
\begin{figure}[t]\centering
\includegraphics[width=0.78\linewidth]{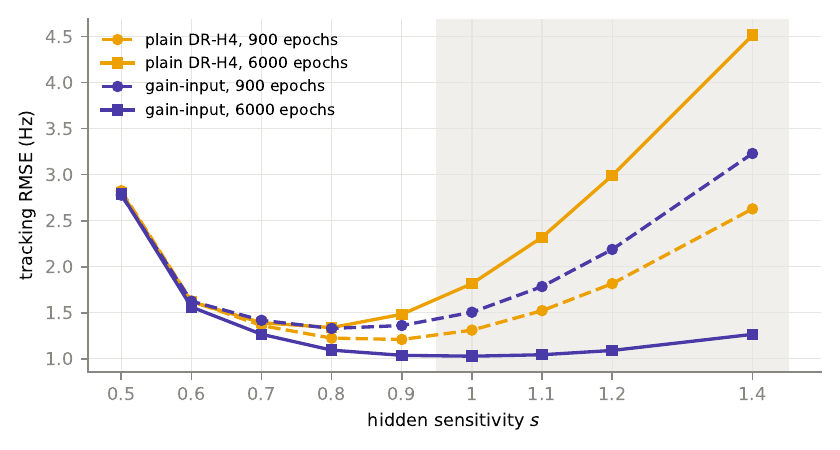}
\caption{Matched-epoch comparison at $\sigma=0.5$: plain DR-H4 and gain-input I2 at 900 (dashed) and 6000 (solid) epochs. Shading marks the high-gain set.}\label{fig:train}
\end{figure}

At matched epochs over the high-gain set, training from 900 to 6000 epochs changed the plain network's RMSE by +1.09 [+1.06, +1.12]\Hz{} and the gain-input network's by $-$1.07 [$-$1.09, $-$1.05]\Hz{} (97.5\% intervals). The directions are opposite, as pre-specified, and the difference-in-differences was 2.16\Hz{} [2.13, 2.20] (Figure~\ref{fig:train}). Given the gain estimate, more training improves both prediction and control; without it, more training improves prediction and degrades control.

\subsection{Secondary comparisons}
Table~\ref{tab:pairs} gives per-sensitivity paired contrasts. I2 and full adaptive ARX differed by between $-$0.037 and +0.052\Hz{}: I2 was better at $s=$ 0.6 and worse at $s=$ 0.9, 1.0, 1.1, 1.2, with no interval-excluding difference elsewhere. No equivalence margin was pre-specified, so we claim closeness, not equivalence. I2 beat anti-windup PI at 9 of 9 sensitivities, by 0.025--0.273\Hz. Closed-loop early stopping (I1) improved on H4-10k for $s\ge$ 0.7 and on-policy data (I3) at 9 of 9 sensitivities, but by much less than I2 at high gain (2.45 and 2.21\Hz{} at $s=1.4$, against 3.21\Hz{} for I2); I1 was slightly worse at $s=0.5$ ($-$0.030 [$-$0.038, $-$0.021]\Hz).

\begin{table}[t]\centering\footnotesize
\caption{Protocol v4 secondary contrasts at $\sigma=0.5$ (Hz; 99.44\% intervals). For I2 $-$ X, negative favours I2; for H4-10k $-$ X, positive favours X.}\label{tab:pairs}
\begin{tabular}{@{}lllll@{}}\toprule
$s$ & I2 $-$ ARX+RLS & I2 $-$ PI+AW & H4-10k $-$ I1 & H4-10k $-$ I3\\\midrule
0.50 & $-$0.015 [$-$0.040, +0.007] & $-$0.081 [$-$0.096, $-$0.067] & $-$0.030 [$-$0.038, $-$0.021] & +0.014 [+0.006, +0.022] \\
0.60 & $-$0.037 [$-$0.080, $-$0.009] & $-$0.097 [$-$0.113, $-$0.078] & $-$0.009 [$-$0.023, +0.006] & +0.043 [+0.031, +0.056] \\
0.70 & $-$0.021 [$-$0.056, +0.004] & $-$0.048 [$-$0.065, $-$0.031] & +0.058 [+0.044, +0.070] & +0.099 [+0.085, +0.112] \\
0.80 & +0.008 [$-$0.012, +0.028] & $-$0.036 [$-$0.051, $-$0.021] & +0.191 [+0.171, +0.214] & +0.218 [+0.202, +0.236] \\
0.90 & +0.052 [+0.025, +0.076] & $-$0.025 [$-$0.047, $-$0.003] & +0.391 [+0.364, +0.431] & +0.389 [+0.360, +0.423] \\
1.00 & +0.043 [+0.009, +0.066] & $-$0.038 [$-$0.055, $-$0.020] & +0.712 [+0.675, +0.748] & +0.657 [+0.622, +0.690] \\
1.10 & +0.042 [+0.004, +0.072] & $-$0.086 [$-$0.112, $-$0.065] & +1.073 [+1.031, +1.119] & +0.973 [+0.932, +1.014] \\
1.20 & +0.024 [+0.005, +0.043] & $-$0.136 [$-$0.155, $-$0.116] & +1.474 [+1.411, +1.546] & +1.311 [+1.251, +1.380] \\
1.40 & +0.008 [$-$0.009, +0.028] & $-$0.273 [$-$0.304, $-$0.247] & +2.449 [+2.392, +2.517] & +2.212 [+2.155, +2.270] \\
\bottomrule\end{tabular}\end{table}

\subsection{Robustness at high process noise}
At $\sigma=2.0$, Q1 was 0.24 [0.17, 0.32]: gain-only adaptation recovered only about a quarter of the full-RLS benefit, well below the threshold used at $\sigma=0.5$. Q3 and Q4 stayed positive but were small (0.042 and 0.018\Hz). At this noise level the predictive arms lay within about 0.15\Hz{} of each other on average (mean RMSE 2.201\Hz{} for adaptive ARX, 2.224 for I2, 2.272 for H4-10k), and anti-windup PI was worse (2.439\Hz). The mechanism we confirmed at $\sigma=0.5$ therefore matters little when process noise dominates the error.

\section{Discussion}
\paragraph{Information and adaptation, not nonlinear capacity.} Removing the history substantially worsened the learned predictor at low and moderate gain ($s=0.6$, $0.8$; at $s\ge1.0$ the memoryless network was better); a linear model with the same history performed as well at low gain; and adapting that linear model online made it the best or near-best controller across almost the entire range in both protocols. At $\sigma=0.5$, protocol v4 indicates that most of what the adaptation provides is an estimate of one multiplicative input gain (this was not so at $\sigma=2.0$). A network trained across randomised gains, but never told the gain, learned an action sensitivity that changed far less than the plant gain did, so its errors grew steeply as the gain increased. Supplying the gain estimate fixed this, which is the explicit identification strategy of universal policies with online system identification~\cite{yu2017} applied to neural-population rate control.

\paragraph{Why better prediction gave worse control.} Without the gain, a network can lower average one-step error by fitting the open-loop data more closely while its action sensitivity remains wrong for any particular plant; the one-step controller then over- or under-acts. With the gain available, lower prediction error corresponds to a more accurate model of the current plant, and control improves. This account fits both confirmed results but rests on a single plant family and a single architecture; we offer it as an interpretation, consistent with objective mismatch in model-based reinforcement learning~\cite{lambert2020,grimm2020}, not as a general law.

\paragraph{Classical feedback is a hard baseline.} The apparent low-gain advantage of the learned controller in the earlier phase largely disappeared once PI was tuned on the same development data and given anti-windup. Claims that a learned controller beats classical control should state how the classical baseline was tuned~\cite{kunapuli2025}. Our results differ from the MPC-over-PID finding in spiking networks~\cite{fehrman2024}; candidate reasons are our one-step action selection, our tuned PI with anti-windup and our fully observed population rate. Among the learned predictors, only the gain-input network beat anti-windup PI at every sensitivity (full adaptive ARX also had lower mean RMSE than anti-windup PI at every sensitivity in v4).

\section{Limitations}
\begin{itemize}\setlength\itemsep{2pt}
\item \textbf{Synthetic plant.} Uncoupled LIF-like neurons with a phenomenological light gate; no recurrence, inhibition, plasticity, spatial light distribution or opsin kinetics. One plant family supplies training and test plants, and the shift is a scalar gain plus process noise, which suits a one-parameter estimator.
\item \textbf{Noise dependence.} Gain-only adaptation recovers most of the adaptive-ARX benefit at $\sigma=0.5$ but little at $\sigma=2.0$, where Q1 fell to 0.24.
\item \textbf{Controller scope.} One-step action selection only; PI from one family tuned by grid search; one learned architecture. The action-sensitivity analysis uses the gain-RLS estimate as its reference, not the true local gain.
\item \textbf{Hypothesis origin.} All v4 hypotheses and the v3 training-length comparison were formed from development data; their confirmation on untouched plants is what we report.
\item \textbf{Run integrity.} Two v3 chunks ran concurrently by mistake and three v4 chunks were interrupted by infrastructure errors; affected plant-conditions were recomputed and matched exactly (Appendix~\ref{app:integrity}).
\end{itemize}

\section{Conclusion}
In a first prospective confirmation on untouched synthetic plants, removing the input--output history substantially worsened the trained predictor at low and moderate gain ($s=0.6$, $0.8$), and a linear model with the same history was practically equivalent at low gain (primary hypotheses); in pre-specified secondary and benchmark comparisons, online linear adaptation outperformed it, anti-windup PI matched or beat it from moderate gain upwards, and longer training improved its predictions while worsening its control. A second protocol, with these observations turned into primary hypotheses, indicated that, in this synthetic plant at the primary noise level ($\sigma=0.5$), most of this pattern follows from one missing quantity: re-estimating a single input gain recovered most of the benefit of adaptive ARX, and giving the network that estimate removed both its high-gain failure and its prediction--control mismatch at the primary noise level. In this synthetic plant at low process noise, explicit estimation of the hidden parameter mattered more than model expressiveness; at $\sigma=2.0$ these effects weakened substantially (gain-only recovery fell to 24\%), which limits the scope of the proposed mechanism to the tested noise regime.

\section*{Code and data availability}
\begin{sloppypar}
Code, frozen protocols (\texttt{PROTOCOL\_V3.json}, \texttt{PROTOCOL\_V4.json}), manifests, model weights, raw per-plant results, development records and the scripts that generate every number and figure in this paper: \url{https://github.com/lonewolf15116/NeuroLight-AI}, archived at \url{https://doi.org/10.5281/zenodo.23263620}.
\end{sloppypar}

\appendix
\section{Study history}\label{app:history}
\begin{table}[h]\centering\small
\begin{tabular}{@{}llL{7.6cm}@{}}\toprule
Stage & Plants & Outcome\\\midrule
v0.1--v0.6 (exploratory) & 0--69, 1000--1049, 1400--1649 & learned controller brittle to gain shift; history helps\\
C0 confirmation & 3000--3049 & chronology effect +2.1 / +5.4\Hz\\
Experiment A & 4000--4049 & matched history ablation +2.76\Hz{} at $s=0.6$\\
Experiment B & 6000--6049 & linear equivalence at $s=0.6$; PI later found under-tuned, RLS inactive\\
Phase-2 development & 5000--5149 & PI search, adaptive RLS, 10\,000-epoch retraining, rehearsal\\
\textbf{v3 confirmation} & 7000--7049 & Section~4\\
Phase-4 development & 5200--5299 & gain mechanism, checkpoints, interventions (Section 5)\\
\textbf{v4 confirmation} & 8000--8049 & Section 6\\\bottomrule
\end{tabular}\end{table}
Phase-1 results are retained as historical records and are not pooled with v3 or v4. In phase 1, PI's integral gain lay on the edge of its tuning grid and switching RLS updates off changed RMSE by less than 0.01\Hz; both were corrected on development plants before v3.

\section{Run integrity}\label{app:integrity}
\textbf{v3.} The confirmation ran in nine resumable invocations on a laptop. Two were accidentally launched concurrently and wrote the same progress file. The simulator is deterministic per plant and each write is a complete snapshot, so the only possible effect is duplicated work; all 200 plant-conditions in the four affected conditions were recomputed from the frozen code, with a maximum absolute difference of 0.0.

\textbf{v4.} The confirmation ran in resumable chunks of at most 172\,s, one at a time. Three invocations returned infrastructure errors with unknown execution status (at about 465, 561 and 869 of 900 plant-conditions); each resume waited until the progress file had been unchanged for over three minutes. Ten plant-conditions (those on both sides of every interruption, plus the first and last primary plant) were recomputed from scratch in a separate environment; all 302 recorded values of each matched exactly. Recomputing every statistic from the stored raw data reproduced the reported values to $3\times10^{-16}$. Results file SHA-256: \texttt{\detokenize{422b4078c07f8136bcb887469c1962ac24cf596a3d6dded5addb3a289e825b5c}}.

\end{document}